\documentclass[twocolumn,english]{article}
\usepackage[T1]{fontenc}
\usepackage[latin9]{inputenc}
\usepackage{babel}
\usepackage{array}
\usepackage{mathrsfs}
\usepackage{mathtools}
\usepackage{multirow}
\usepackage{amstext}
\usepackage{amssymb}
\usepackage{graphicx}
\usepackage{wasysym}
\usepackage[unicode=true,pdfusetitle,
 bookmarks=true,bookmarksnumbered=false,bookmarksopen=false,
 breaklinks=false,pdfborder={0 0 1},backref=false,colorlinks=false]
 {hyperref}

\makeatletter

\providecommand{\tabularnewline}{\\}
\newcommand{\lyxdot}{.}

\@ifundefined{date}{}{\date{}}
\@ifundefined{showcaptionsetup}{}{%
 \PassOptionsToPackage{caption=false}{subfig}}
\usepackage{subfig}
\makeatother

\begin{document}
\title{CourtNet for Infrared Small-Target Detection}
\author{Jingchao Peng, Haitao Zhao, Kaijie Zhao, Zhongze Wang, and Lujian Yao\\
Automation Department, School of Information Science and Engineering\\
East China University of Science and Technology}
\maketitle
\begin{abstract}
Infrared small-target detection (ISTD) is an important computer vision
task. ISTD aims at separating small targets from complex background
clutter. The infrared radiation decays over distances, making the
targets highly dim and prone to confusion with the background clutter,
which makes the detector challenging to balance the precision and
recall rate. To deal with this difficulty, this paper proposes a neural-network-based
ISTD method called CourtNet, which has three sub-networks: the prosecution
network is designed for improving the recall rate; the defendant network
is devoted to increasing the precision rate; the jury network weights
their results to adaptively balance the precision and recall rate.
Furthermore, the prosecution network utilizes a densely connected
transformer structure, which can prevent small targets from disappearing
in the network forward propagation. In addition, a fine-grained attention
module is adopted to accurately locate the small targets. Experimental
results show that CourtNet achieves the best F1-score on the two ISTD
datasets, MFIRST (0.62) and SIRST (0.73).
\end{abstract}

\section{Introduction}

Target detection is one of the key technology in the area of computer
vision \cite{SSD,FasterRCNN,YOLO}. With the development of deep learning,
target detection makes considerable progress, and has become a crucial
component of intelligent robotics, autonomous driving applications,
smart cities, etc. Among them, infrared small-target detection (ISTD)
attracts more and more attention because of the great ability to detect
at degraded visibility conditions and a great distance. Compared with
visible light imaging, infrared imaging offers strong anti-interference
and ultra-long-distance detection capability \cite{survey}. When
the target is far from the infrared detector, it usually occupies
a small region, and the target is easily submerged in background clutter
and sensor noises. Because of two main characteristics, the small
size and the complex background clutter, ISTD is still a big challenge
in the field of target detection \cite{TSUCAN,MDvsFAcGAN}.

\begin{figure}[!t]
\begin{centering}
\includegraphics[width=0.45\textwidth]{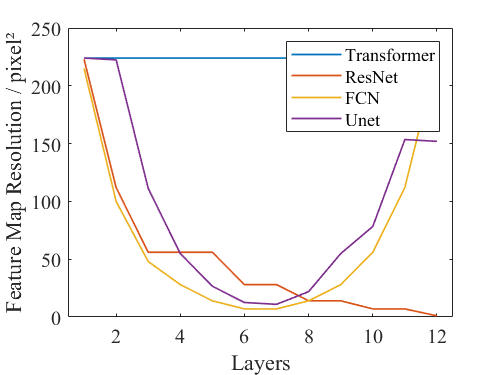}
\par\end{centering}
\caption{\label{fig:resolution}The difference in feature map resolution between different networks.}
\end{figure}

The first characteristic of ISTD is the small size of the targets.
The Society of Photo-Optical Instrumentation Engineers (SPIE) describes a small target as follows: the target size is usually less than 9$\times$9 pixels, and the area proportion is less than 0.15\% \cite{smallsize}.
currently, most of ISTD methods use deep convolution neural networks (DCNNs) to detect small targets.
However, the resolution of feature maps of DCNN is gradually reduced in forward propagation, as shown in Fig. \ref{fig:resolution}.
In this case, such a small target is easy to disappear in forward propagation \cite{DRPN}.
Many DCNN-dased methods simplify network architectures to alleviate the problem of small targets disappearing \cite{purning1,purning2,purning3}.
But this simplification will bring another problem of insufficient expressive capacity \cite{complexity}, say insufficient model complexity resulting from insufficient parameters, especially under complex backgrounds.

Transformers, however, are quite different.
The feature resolution of Transformers remains unchanged with the network going from shallow to deep, as shown in Fig. \ref{fig:resolution}, which is naturally beneficial for ISTD.
Current state-of-the-art Vision Transformers (ViT) have surpassed the performance of DCNNs in many fields \cite{VIT}.
However, the original attention module (called coarse-grained attention in this paper, compared to the proposed fine-grained attention) in ViT is performed at the patch level, whose size is 16$\times$16 (much larger than the target size).
In other words, transformers can select the patch where the target is, but cannot accurately locate the target in the patch.
Moreover, the original features of infrared images will vanish after multiple blocks on the deep network \cite{DenseFormer}.
The original features contain the location and size information of the targets, which is crucial to ISTD.

The second characteristic of ISTD is the complex background clutter.
Since the essence of infrared images is to receive thermal radiation,
given that the infrared radiation energy of false alarm sources such
as thermal light sources, cirrus clouds, and edges and corners of
the buildings are similar to that of targets, there will inevitably
be many noises in the infrared images \cite{noise}. The size of the
noises is similar to that of targets, so the detector is easy to regard
noises as targets, resulting in a low precision rate of the detector.
If we adjust the detector so that it is less sensitive to noises,
it will lead to the miss detection of real targets, resulting in a
low recall rate of the detector. Therefore, it is important for the
detector to balance the precision rate and the recall rate \cite{TSUCAN,MDvsFAcGAN}. 

The existing ensemble methods combine two or more models \cite{TSUCAN,MDvsFAcGAN,AVILNet},
which respectively focus on the precision rate and the recall rate.
But the outputs of the two detectors are averaged, so that when one
of the detectors thinks the target is real but the other detector
thinks the target is fake, these methods still need to manually adjust
the threshold to balance the precision rate and the recall rate.

\begin{figure}[!t]
\begin{centering}
\includegraphics[width=0.75\columnwidth]{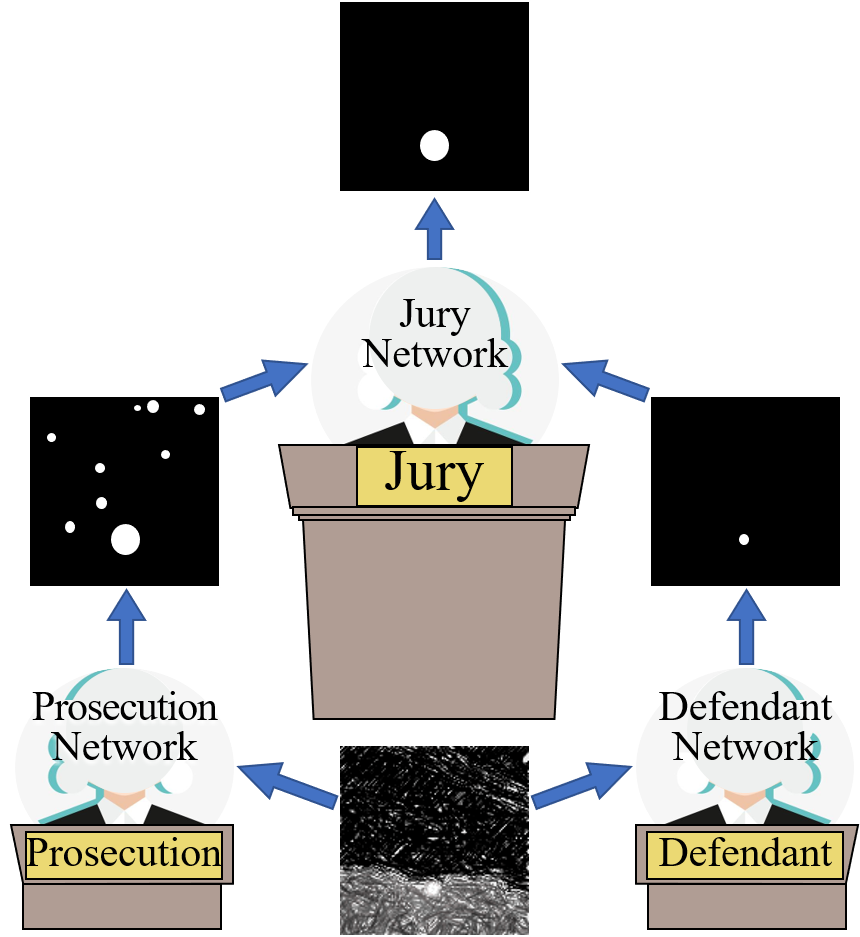}
\par\end{centering}
\caption{\label{fig:The-overall-structure}The overall structure of CourtNet.
The prosecution network finds all targets; the defendant network filters
noise; the jury network makes the final decision.}
\end{figure}

Motivated by the above analysis, in this paper, we propose a novel
ISTD method called CourtNet, as shown in Fig. \ref{fig:The-overall-structure}.
CourtNet borrows ideas from court debate, consisting of a prosecution
network, a defendant network, and a jury network. The prosecution
network finds all suspected targets as much as possible to improve
the recall rate; in contrast, the defendant network is quite conservative
and avoids considering noise as a target to increase the precision
rate. Their outputs are adjudicated by the jury network. The jury
network takes the weighted sum of their results to get the final result.
So that CourtNet can achieve an adaptive balance of the precision
and recall rate. In addition, we propose an adaptive balance loss,
which adds an adaptive scaling factor to the loss item representing
the precision rate or the recall rate. So that during the training
phase, CourtNet can adaptively enhance the precision rate or the recall
rate according to a larger loss value. Experiments show that the adaptive
balance loss can reduce fluctuations and accelerate convergence.

For the consideration of small targets, a densely connected transformer
is proposed as the prosecution network. the densely connected transformer
improves the ViT by adopting a dense connection, which can prevent
small targets from disappearing in the forward propagation by retaining
shallow detail features. In addition to the coarse-grained attention
module in ViT, the prosecution network performs fine-grained attention
inside each patch to locate the target at pixel level.

In summary, our contributions are summarized below:
\begin{enumerate}
\item CourtNet adopts a prosecution-defendant-jury network structure. The
prosecution network focuses on improving the recall rate; the defendant
network focuses on improving the precision rate; the jury network
makes the final decision, adaptively balancing the precision rate
and the recall rate.
\item A densely connected transformer is proposed as the prosecution network
to prevent small targets from disappearing in the forward propagation
by retaining shallow detail features. Complemented with a coarse-grained
attention module, the prosecution network performs a fine-grained
attention module inside each patch to accurately locate small targets.
\item CourtNet adopts an adaptive balance loss, which can adaptively enhance
the loss item which represents the precision rate or the recall rate
according to their loss values.
\end{enumerate}
\begin{figure*}[t]
\begin{centering}
\subfloat[Illustration of the prosecution network.]{\begin{centering}
\includegraphics[width=0.85\textwidth]{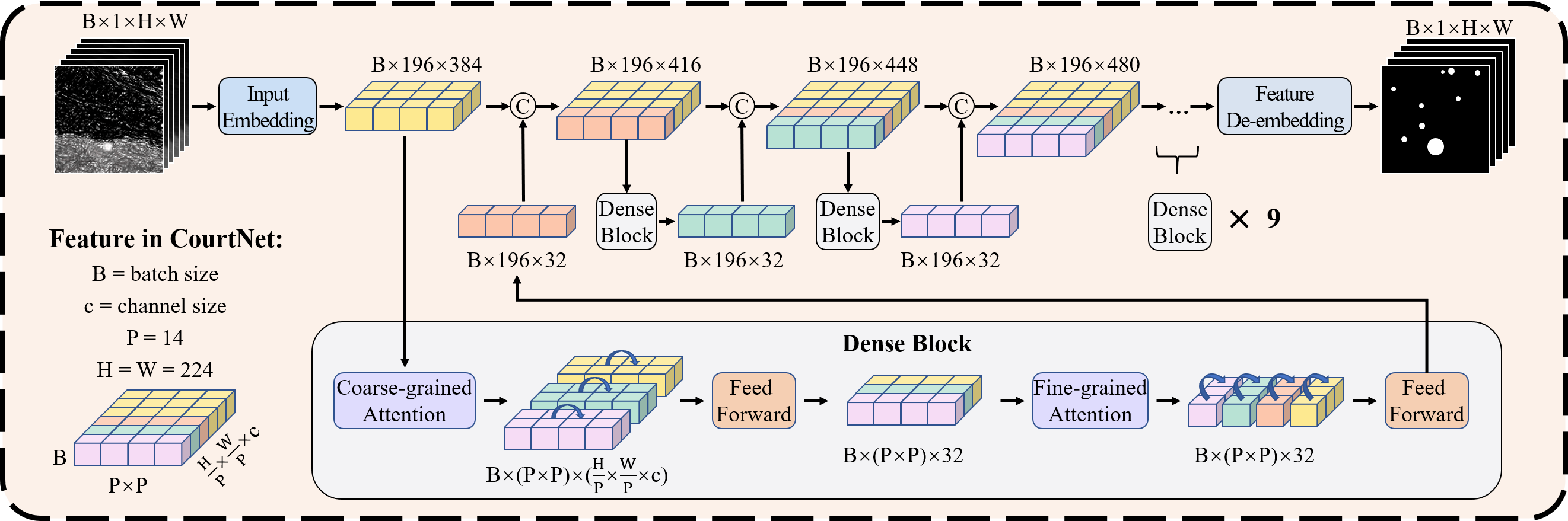}
\par\end{centering}
}
\par\end{centering}
\begin{centering}
\subfloat[Illustration of the defendant network.]{\begin{centering}
\includegraphics[width=0.85\textwidth]{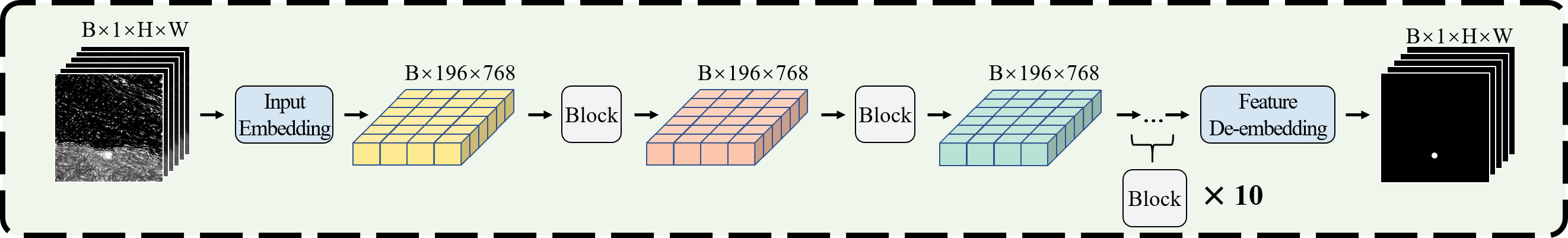}
\par\end{centering}
}
\par\end{centering}
\begin{centering}
\subfloat[Illustration of the jury network.]{\begin{centering}
\includegraphics[width=0.85\textwidth]{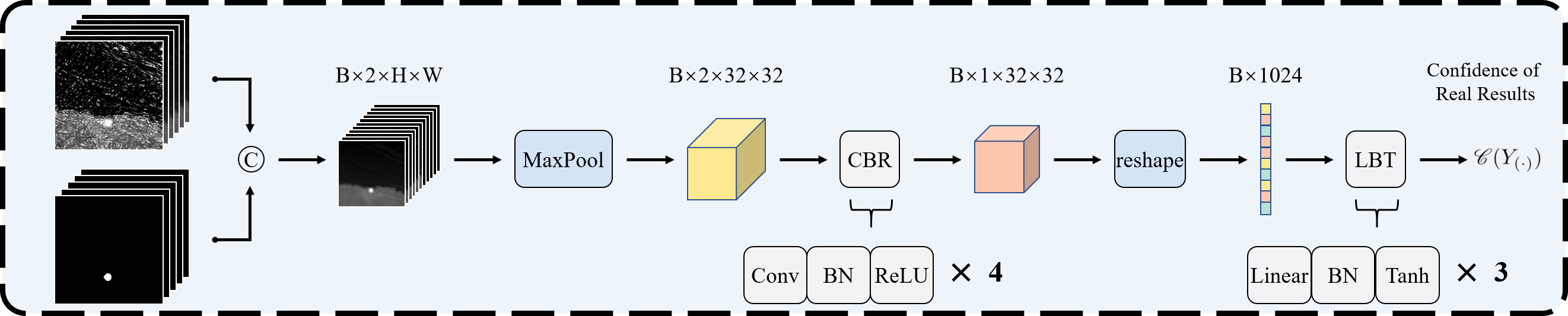}
\par\end{centering}
}
\par\end{centering}
\caption{The specific structure of CourtNet.\label{fig:The-specific-structure}}
\end{figure*}

\section{Related Works}

ISTD methods can largely be divided into traditional methods and deep
learning-based methods. Traditional ISTD methods are based on the
contrast mechanism of the human visual system, such as DoG \cite{DoG}
and LCM \cite{LCM}. WSLCM \cite{WSLCM} utilized a weighting function
and strengthened LCM to consider characters and differences between
targets and backgrounds. TLLCM \cite{TLLCM} adopted multi-scale multi-layer
LCM to deal with infrared small targets of different types and sizes.
Considering that the gray values of small targets are usually higher
than that of the background, AADCDD \cite{ADDCDD} took advantage
of local gray differences and employed weighting coefficients to detect
targets. ADMD \cite{ADMD} utilized directional information to detect
targets in high gray-intensity structural backgrounds. MSPCM \cite{MSPCM}
aggregated multi algorithms to increase the precision rate. These
traditional methods work well in simple backgrounds but are disturbed
by complex background clutter \cite{survey}.

Deep-learning-based methods, such as SSD \cite{SSD}, Faster R-CNN
\cite{FasterRCNN}, YOLO \cite{YOLO}, U-Net \cite{UNet}, and so
on, achieved excellent results in visible light object detection and
segmentation. In order to deal with ISTD, many technologies has been
proposed, including model pruning \cite{purning1,purning2,purning3},
multi-scale fusion \cite{MSfusion1,MSfusion2,MSfusion3}, multi-modal
fusion \cite{MMfusion1,MMfusion2,MMfusion3}, feature pyramid \cite{FPN1,FPN2,FPN3},
etc. ACM \cite{ACM} and ALCNet \cite{ALCNet} utilized a top-down
global attention module and a bottom-up local attention module to
separately transfer semantic information and context information,
and to prevent the disappearance of small targets. However, due to
the feature resolution of convolutional networks gradually decreasing,
features of small infrared targets can hardly be preserved. As for
balancing the precision rate and the recall rate, TS-UCAN \cite{TSUCAN}
utilized two sub-networks in series connection, the first sub-network
improves the recall rate, and the second sub-network improves the
precision rate. MDvsFA-cGAN \cite{MDvsFAcGAN} and AVILNet \cite{AVILNet}
utilized multi sub-networks in parallel connection and cooperate with
the generative adversarial network (GAN). But they still need to manually
adjust the threshold while cannot adaptively balance the precision
rate and the recall rate.

\begin{figure*}[t]
\begin{centering}
\subfloat[Coarse-grained attention]{\begin{centering}
\includegraphics[width=0.85\columnwidth]{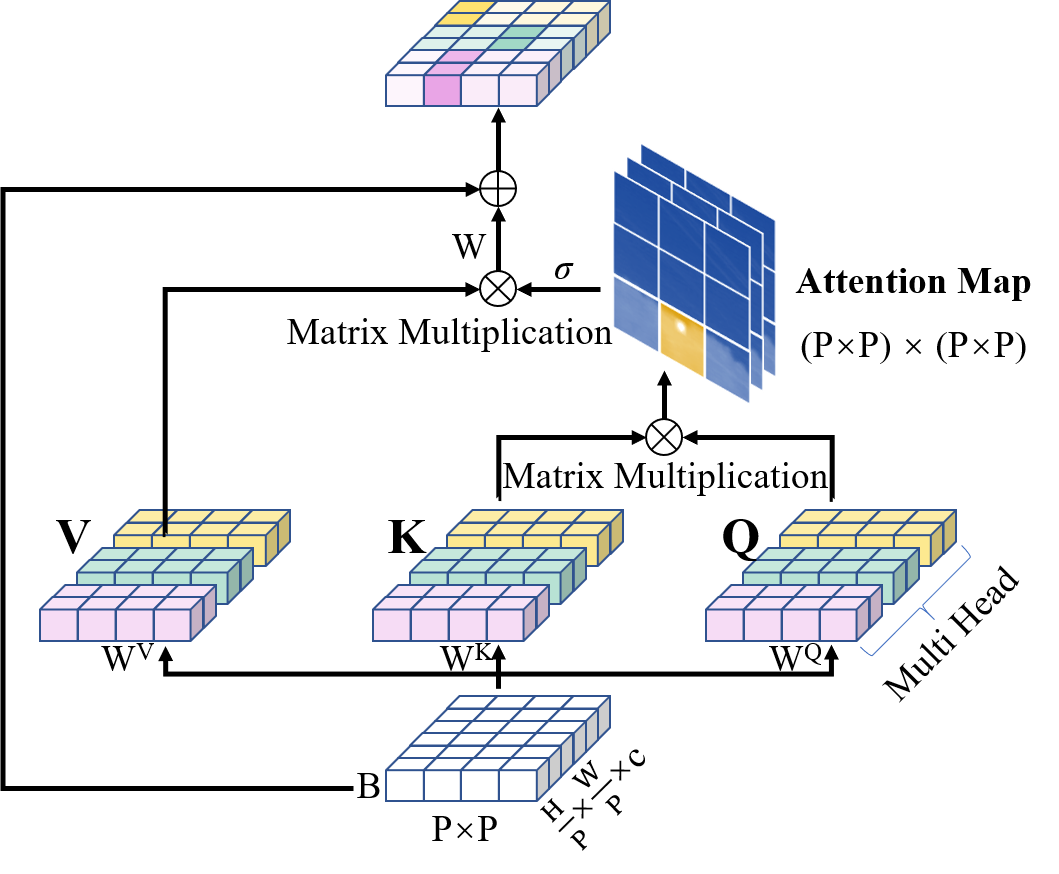}
\par\end{centering}
}\ \ \ \ \ \ \ \ \ \ \ \ \ \ \ \subfloat[Fine-grained attention]{\begin{centering}
\includegraphics[width=0.85\columnwidth]{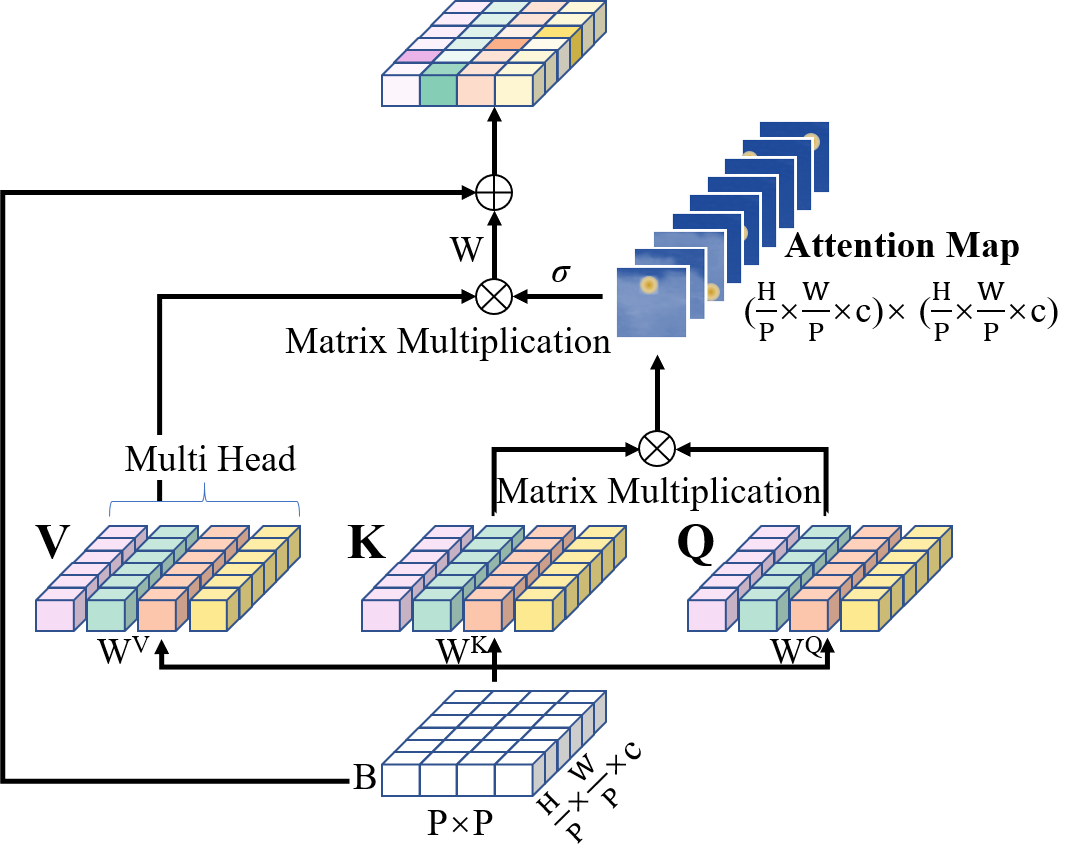}
\par\end{centering}
}
\par\end{centering}
\caption{\label{fig:Comparison-of-the-1}Comparison of the coarse-grained attention
and the fine-grained attention.}
\end{figure*}

\section{Proposed Method}

In this section, we first overview the architecture of CourtNet. Then
we introduce the proposed densely connected transformer and the fine-grained
attention module in the denseblock. Finally, we introduce the adaptive
balance loss as well as other losses utilized for the training phase.

\subsection{Overall Structure}

CourtNet consists of a prosecution network, a defendant network, and
a jury network, as shown in Fig. \ref{fig:The-overall-structure}.
Assuming that the infrared image is a suspect, the task of the prosecution
network is to find out all targets (crimes), which requires the network
to be sensitive to the details and not to miss any targets. The specific
details of the prosecution network can be seen in next section (Densely
Connected Transformer). The output of the prosecution network is $Y_{P}=\mathscr{P}(X_{0})$,
where $X_{0}$ is the input, $\mathscr{P}(\cdot)$ is the mapping
of the prosecution network.

The task of the defendant network is to fully \textit{defend} the
infrared image, which requires the network to be robust to false positive
targets. Since the features are fed forward through the blocks to
generate new features, the original features containing noises will
vanish after multiple blocks, therefore ViT can filter parts of noises.
So we use ViT as the defendant network. The structure of the defendant
network is shown in Fig. \ref{fig:The-specific-structure} (b), which
consists of an input embedding module, 12 blocks, and a feature de-embedding
module. The outputs of the defendant network is $Y_{D}=\mathscr{D}(X_{0})$,
where $X_{0}$ is the input, $\mathscr{D}(\cdot)$ is the mapping
of the defendant network.

The task of the jury network is to judge whether the results $Y_{(\cdot)}$
from the prosecution network or the defendant network are true, its
output is the confidence $\mathscr{C}(Y_{(\cdot)})\in[0,1]$. Inspired
by Wang et al. \cite{MDvsFAcGAN}, we adopt a convolution network
as the jury network. The structure of the jury network is shown in
Fig. \ref{fig:The-specific-structure} (c), the jury network consists
of 4 convolution layers and 3 fully connected layers. The final result
$Y$ is the weighted sum of the results from the prosecution network
and the defendant network: {\small{}
\begin{equation}
Y=w_{p}\times Y_{P}+w_{d}\times Y_{D},
\end{equation}
where $w_{p}=\frac{\mathscr{C}(Y_{P})}{\mathscr{C}(Y_{P})+\mathscr{C}(Y_{D})}$,
$w_{d}=\frac{\mathscr{C}(Y_{D})}{\mathscr{C}(Y_{P})+\mathscr{C}(Y_{D})}$,
means the weight of $Y_{P}$ and $Y_{D}$}, respectively{\small{}.}{\small\par}

\subsection{Densely Connected Transformer\label{sec:DCT}}

The prosecution network utilizes a densely connected transformer structure,
which consists of an input embedding module, 12 dense blocks, and
a feature de-embedding module. The structure is shown in Fig. \ref{fig:The-specific-structure}
(a). Consider a batch of infrared images $X_{0}\in\mathbb{R}^{B\times1\times H\times W}$,
The batch size is $B$. Each image has one gray channel and its width
and height are $W$ and $H$, respectively. After the input embedding
module, the image is divided into $p\times p$ patches with width,
height, and channel of $\frac{H}{p}$, $\frac{W}{p}$, and $c$, respectively.
In other words, $X_{0}\in\mathbb{R}^{B\times1\times H\times W}$ are
encoded as the feature $F_{0}\in\mathbb{R}^{B\times P\times C}$,
where $P=p\times p$, $C=\frac{H}{p}\times\frac{W}{p}\times c$. 

Then the feature $F_{0}$ goes through 12 dense blocks to get the
final feature $F_{12}$. Suppose the input of the $i$-th dense block
is $F_{i-1}\in\mathbb{R}^{B\times P\times C_{i-1}}$, the output is
$\mathscr{F}(F_{i-1})\in\mathbb{R}^{B\times P\times32}$, Then the
input feature and output feature are concatenated:
\begin{equation}
F_{i}=\textrm{concat}(F_{i-1},\mathscr{F}(F_{i-1}))\in\mathbb{R}^{B\times P\times C_{i}},
\end{equation}
where $C_{i}=C_{i-1}+32$. For the ViT \cite{VIT}, the input feature
dimension of each block is equal to the output feature dimension,
and the input feature is not preserved.The output feature dimension
of our dense block is 32. The input feature is concatenated with the
output feature. The advantage of concatenation is to 1) reduce the
amount of computation and increase the detection speed, 2) retain
original features, which contain location and size information of
the targets.

Finally, after the feature de-embedding module, outputs are obtained
by Eq. 1. The outputs $Y\in\mathbb{R}^{B\times1\times H\times W}$
are binary maps with 0 and 1 indicating whether each pixel is a target
or not.

\subsection{Denseblock}

The dense block consists of a coarse-grained attention module, a fine-grained
attention module, and two feed-forward modules, as shown in grey box
of Fig. \ref{fig:The-specific-structure} (a). The structures of the
coarse-grained attention module is shown in Fig. \ref{fig:Comparison-of-the-1}
(a). The coarse-grained attention module adopts the form of multi-heads,
the input feature $F\in\mathbb{R}^{P\times C}$ (not considering the
batch here) is divided into $n$ groups, and each group of feature
is $F_{j}\in\mathbb{R}^{P\times\frac{C}{n}},\;j=1,\ldots,n$. The
coarse-grained attention map is calculated by{\small{}
\begin{equation}
\begin{array}{c}
\mathclap{A_{\textrm{coarse}}=\left[\begin{array}{cccc}
\cdots & Q_{j-1}K_{j-1}^{T} & Q_{j}K_{j}^{T} & \cdots\end{array}\right]\in\mathbb{R}^{P\times P}}\\
j=1,\ldots,n,
\end{array}
\end{equation}
}where $Q_{j}$ and $K_{j}$ represent query and key, respectively.
It is worth noting that the coarse-grained attention map $A_{\textrm{coarse}}\in\mathbb{R}^{P\times P}$,
where $P$ is the number of patches. The function of the coarse-grained
attention module is to select patches where the target is located
in by weighting patches.

The structures of the fine-grained attention module is shown in Fig.
\ref{fig:Comparison-of-the-1} (b). The fine-grained attention module
adopts a different grouping method from the coarse-grained attention
module and divided $F$ into $m$ groups, each group of feature is
$F_{k}\in\mathbb{R}^{\frac{P}{m}\times C},\;k=1,\ldots,m$. The fine-grained
attention map is calculated by{\small{}
\begin{equation}
A_{\textrm{fine}}=\left[\begin{array}{c}
\vdots\\
Q_{k-1}^{T}K_{k-1}\\
Q_{k}^{T}K_{k}\\
\vdots
\end{array}\right]\in\mathbb{R}^{C\times C},\;k=1,\ldots,m.
\end{equation}
}Since $C=\frac{H}{p}\times\frac{W}{p}\times c$, this dimension contains
not only channel information but also spatial information inside each
patch, so the function of the fine-grained attention map is to select
and weight position inside each patch.

\subsection{Loss Formulation}

Given the result of CourtNet as $Y$ and the ground truth as $\hat{Y}$,
we set $Pr=\frac{\Sigma(Y\times\hat{Y})}{\Sigma Y}$. The closer $Pr$
is to 1, the more targets detected by CourtNet are real. Therefore
$Pr$ reflects the precision rate. We set $Re=\frac{\Sigma(Y\times\hat{Y})}{\Sigma\hat{Y}}$.
The closer $Re$ is to 1, the more real targets are detected by CourtNet.
Therefore $Re$ reflects the recall rate. Inspired by Lin et al.\cite{focalloss},
we propose an adaptive balance loss:
\begin{equation}
\mathcal{L}_{\textrm{abl}}=-(1-Pr)^{\gamma}\times\log(Pr)-(1-Re)^{\gamma}\times\log(Re),
\end{equation}
where $-(1-\cdot)^{\gamma},\;\gamma\in\mathbb{N^{\ast}}$ means adaptive
scaling factor, here we set $\gamma=3$. The adaptive balance loss
can adaptively enhance the large loss item. For example, When the
precision rate of the current model is high enough, $Pr$ is close
to 1, the value of $(1-Pr)^{\gamma}$ is small, and the loss item
$(1-Pr)^{\gamma}\times\log(Pr)$ has a negligible impact on loss $\mathcal{L}_{\textrm{abl}}$.
On the contrary, when the precision rate of the current model is low,
$Pr$ is close to 0, the value of $(1-Pr)^{\gamma}$ is large, and
the loss item $(1-Pr)^{\gamma}\times\log(Pr)$ has a great impact
on loss $\mathcal{L}_{\textrm{abl}}$.

The loss of the prosecution network and the defendant network is
\begin{equation}
\mathcal{L}_{P}=10\times\mathcal{L}_{\textrm{abl}}-\log(\mathscr{C}(Y_{P})),
\end{equation}
\begin{equation}
\mathcal{L}_{D}=10\times\mathcal{L}_{\textrm{abl}}-\log(\mathscr{C}(Y_{D})),
\end{equation}
where the loss item $\log(\mathscr{C}(Y_{(\cdot)}))$ makes the output
of the prosecution network or the defendant network close to ground
truth.

The jury network adopts cross entropy loss:
\begin{equation}
\mathcal{L}_{J}=-Y^{\ast}\log(\mathscr{C}(Y_{(\cdot)}))-(1-Y^{\ast})\log(1-\mathscr{C}(Y_{(\cdot)})),
\end{equation}
where $Y_{(\cdot)}$ refers to the detection results, if $Y_{(\cdot)}$
is the same as the ground truth, $Y^{\ast}=1$, otherwise $Y^{\ast}=0$.

\section{Experiments}

In this section, we will introduce experimental settings, such as datasets as well as evaluation metrics, and compare CourtNet with other methods. 

\subsection{Experimental Settings}

\begin{table*}[!t]
\caption{\label{tab:Comparison-of-different}Comparison of different methods which were evaluated on MFIRST and SIRST.}
\resizebox{\textwidth}{!}{
\begin{centering}
\begin{tabular*}{1\textwidth}{@{\extracolsep{\fill}}cccccccc}
\hline 
\multirow{2}{*}{Methods} & \multicolumn{3}{c}{MFIRST} & \multicolumn{3}{c}{SIRST} & Time\tabularnewline
\cline{2-7} \cline{3-7} \cline{4-7} \cline{5-7} \cline{6-7} \cline{7-7} 
 & Precision & Recall & F1 & Precision & Recall & F1 & {\footnotesize{}(s per 100 images)}\tabularnewline
\hline 
ADMD \cite{ADMD} & 0.41 & 0.44 & 0.36 & 0.57 & 0.69 & 0.54 & \textbf{0.09}\tabularnewline
MSPCM \cite{MSPCM} & 0.49 & 0.49 & 0.39 & 0.69 & 0.69 & 0.63 & 222.93\tabularnewline
AADCDD \cite{ADDCDD} & 0.50 & 0.56 & 0.42 & 0.79 & 0.63 & 0.64 & 0.19\tabularnewline
TLLCM \cite{TLLCM} & 0.58 & 0.46 & 0.45 & 0.78 & 0.59 & 0.61 & 104.78\tabularnewline
WSLCM \cite{WSLCM} & 0.69 & 0.61 & 0.58 & 0.75 & 0.72 & 0.67 & 162.88\tabularnewline
\hline 
MDvsFA-cGAN \cite{MDvsFAcGAN} & 0.66 & 0.54 & 0.60 & \textbf{0.86} & 0.67 & 0.72 & 0.61\tabularnewline
ALCNet \cite{ALCNet} & \textbf{0.72} & 0.55 & 0.58 & 0.78 & 0.69 & 0.70 & 3,39\tabularnewline
ACM \cite{ACM} & \textbf{0.72} & 0.58 & 0.58 & 0.71 & 0.66 & 0.66 & 1.78\tabularnewline
DNANet\cite{DNANet} & 0.57 & 0.71 & 0.58 & 0.61 & \textbf{0.91} & 0.71 & 1.60 \tabularnewline
CourtNet (Ours) & 0.61 & \textbf{0.75} & \textbf{0.62} & 0.76 & 0.76 & \textbf{0.73} & 1.65\tabularnewline
\hline 
\end{tabular*}
\par\end{centering}
\centering{}}
\end{table*}

We train and test CourtNet on the PyTorch platform with Ryzen 9 5900X CPU and RTX 3090Ti GPU.
First, the prosecution and defendant networks pre-trained by MAE \cite{MAE} are adopted to initialize CourtNet.
The pre-trained dataset is ImageNet \cite{ImageNet}, COCO \cite{COCO}, and FLIR mixed dataset.
The basic learning rate is 5e-4, and the number of epochs is 170.
Then, we utilize Adam as the optimizer to train CourtNet.
The scheduler is warm-up, the warmed-up steps are 200, the beginning learning rate is 1e-7, the max learning rate is 2.5e-5, and the number of epochs is 100.
We use two ISTD datasets, MFIRST \cite{MDvsFAcGAN} and SIRST \cite{ACM}, to evaluate CourtNet.
MFIRST contains 9900 training images and 100 test images, and SIRST contains 341 training images and 86 test images.
As the same as Wang et al.\cite{MDvsFAcGAN}, we use the precision rate, the recall rate, and F1-score to evaluate the binary segmentation result. 
The precision rate, the recall rate, and F1-score are calculated by:
\begin{equation}
Precision=\frac{TP}{TP+FP},
\end{equation}
\begin{equation}
Recall=\frac{TP}{TP+FN},
\end{equation}
\begin{equation}
F1-Score=\frac{2\times Precision\times Recall}{Precision+Recall}.
\end{equation}

\subsection{Comparison with Other Methods}

We compare CourtNet with traditional methods (WSLCM \cite{WSLCM}, TLLCM \cite{TLLCM}, ADMD \cite{ADMD}, MSPCM \cite{MSPCM}, and AADCDD \cite{ADDCDD}) and deep-learning-based methods (MDvsFA-cGAN \cite{MDvsFAcGAN}, ALCNet \cite{ALCNet}, ACM \cite{ACM}, and DNANet\cite{DNANet}).
According to \cite{MDvsFAcGAN}, only achieving a high precision rate or high recall rate does not necessarily indicate good performance, F-measure shall be the primary evaluation.
CourtNet gets the best results for F1-score in two datasets over other methods, as shown in Tab. \ref{tab:Comparison-of-different}.
Specifically, CourtNet achieves 0.62, and 0.73 for F1-score, which is 0.2 and 0.1 higher than the second-best method (MDvsFA-cGAN).
Besides, CourtNet achieves the best results in the recall rate (0.75 for MFIRST, 0.76 for SIRST), which demonstrate the effectiveness of the prosecution network.
Among deep-learning-based methods, CourtNet has the closest precision rate and recall rate, indicating the effectiveness of the proposed prosecution-defendant-jury network structure in balancing the precision rate and the recall rate.

In terms of detecting speed, although CourtNet integrates three networks, it takes 1.65 seconds for CourtNet to detect 100 images, which is faster than five methods (traditional methods: WSLCM, ILCM, and MSPCM; deep-learning-based methods: ALCNet and ACM).
This is because: 1) the prosecution network and the defendant network are independent and can parallel compute, 2) the jury network has a simple structure and light computing burden.

\subsection{Ablation Study}

\subsubsection{Contributions of Different Modules}

\begin{table*}[!h]
\caption{\label{tab:Ablation-study-on}Ablation study on the MFIRST dataset.
''$\checked$'' indicates using the corresponding module, while ''$\times$''
indicates not. ''$-$'' means not suitable.}
\resizebox{\textwidth}{!}{
\begin{tabular}{cccccccccc}
\hline 
\multirow{2}{*}{} & \multirow{2}{*}{Methods} & Dense & Fine-Grained & Jury & \multirow{2}{*}{Precision} & \multirow{2}{*}{Recall} & \multirow{2}{*}{F1} & Flops & Params\tabularnewline
 &  & Connection & Attention & Network &  &  &  & (GMac) & (M)\tabularnewline
\hline 
\multirow{3}{*}{Single} & ViT & $\times$ & $\times$ & $-$ & 0.58 & 0.72 & 0.59 & 16.86 & 86.57\tabularnewline
 & PNet\_v1 & $\checked$ & $\times$ & $-$ & 0.60 & 0.61 & 0.55 & 6.43 & 32.79\tabularnewline
 & PNet & $\checked$ & $\checked$ & $-$ & 0.56 & 0.75 & 0.59 & 6.79 & 60.48\tabularnewline
\hline 
\multirow{2}{*}{Ensemble} & CourtNet\_v1 & $\checked$ & $\checked$ & $\times$ & 0.63 & 0.67 & 0.58 & 23.65 & 147.05\tabularnewline
 & CourtNet & $\checked$ & $\checked$ & $\checked$ & 0.61 & 0.75 & 0.62 & 23.66 & 147.15\tabularnewline
\hline 
\end{tabular}}
\end{table*}

\begin{figure}[!h]
\begin{centering}
\subfloat[Comparison loss value of training set.]{\includegraphics[width=0.39\textwidth]{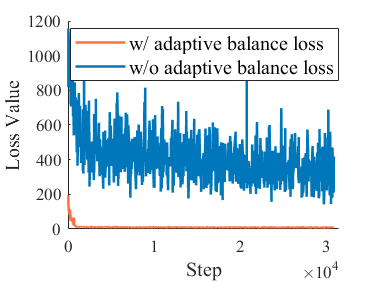}}
\par\end{centering}
\begin{centering}
\subfloat[Comparison F1-score of test set.]{\includegraphics[width=0.39\textwidth]{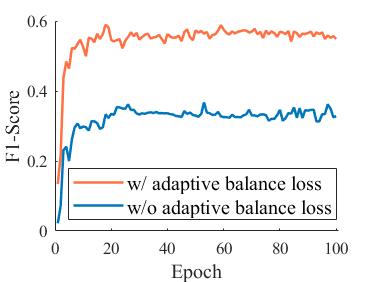}}
\par\end{centering}
\caption{\label{fig:Comparison-of-the}Comparison of the training process w/ adaptive balance loss and w/o adaptive balance loss.}
\end{figure}

\begin{table}[!t]
\centering
\caption{\label{tab:gamma}Ablation study on $\gamma $.}
\begin{centering}
\begin{tabular}{cccc}
\hline
$\gamma $ & Precision & Recall & F1-score \\ \hline
0 & 0.41      & 0.45   & 0.37 \\
1 & 0.65      & 0.45   & 0.47 \\
2 & 0.59      & 0.62   & 0.54 \\
3 & 0.56      & 0.75   & 0.59 \\
4 & 0.57      & 0.70   & 0.57 \\
5 & 0.53      & 0.70   & 0.55 \\ \hline
\end{tabular}
\end{centering}
\end{table}

\begin{figure*}[!h]
\begin{centering}
\subfloat[The prosecution network focuses on the recall rate.]{\includegraphics[width=0.5\textwidth]{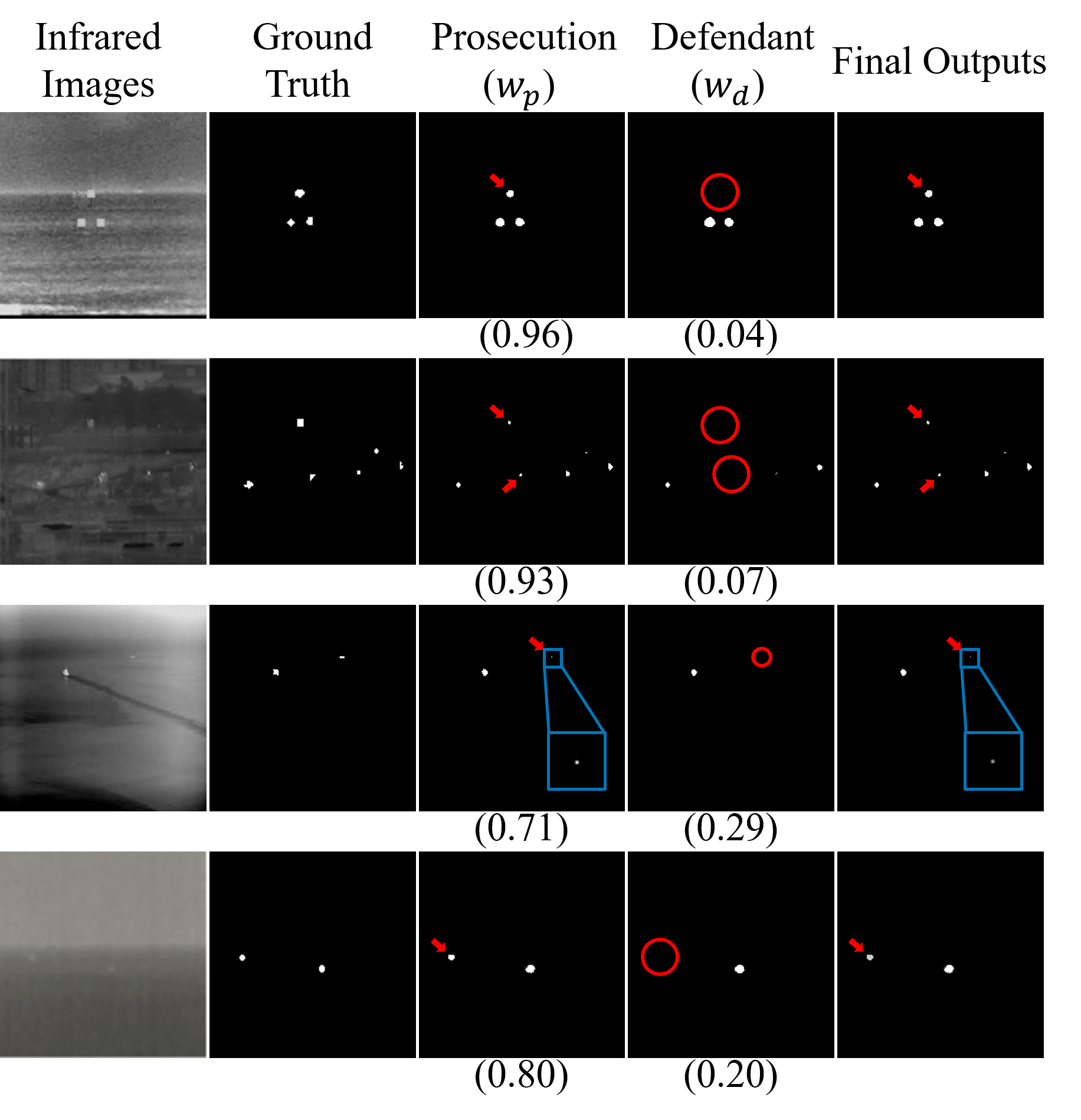}}
\subfloat[The defendant network focuses on the precision rate.]{\includegraphics[width=0.5\textwidth]{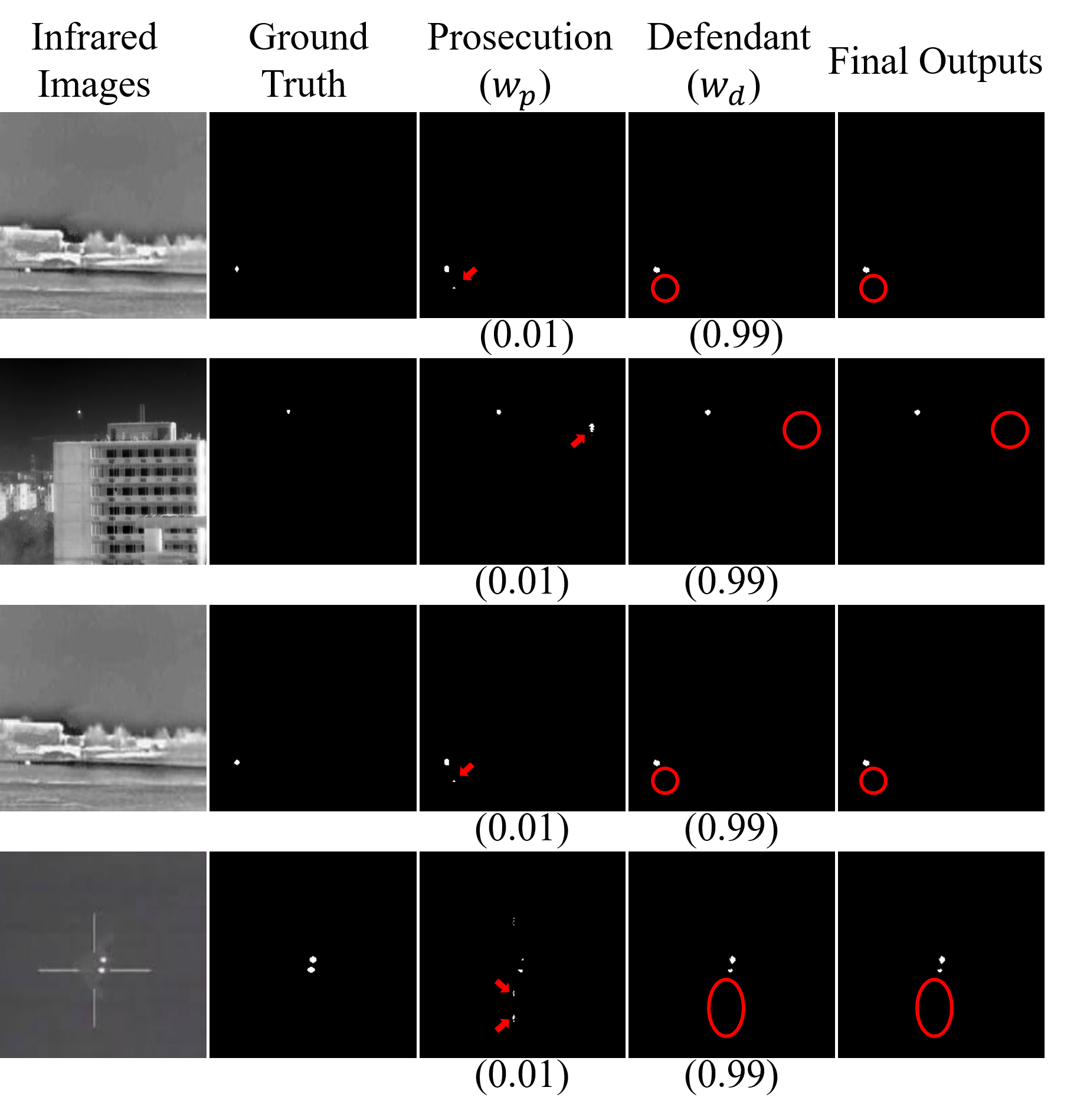}}
\par\end{centering}
\caption{\label{fig:Qualitative-performances-of}Qualitative performances of the prosecution-defendant-jury network structure.
The red arrow indicates that the targets are detected (Positive), and the red circle means the network does not detect any target in this area (Negative).}
\end{figure*}

In order to demonstrate the effectiveness of our major contributions, we implement three variations, including 1) pruning the dense connection; 2) pruning the fine-grained module; 3) pruning the jury network.
Among them, the first two variations adopt a single model to detect, which cannot add the jury network.
The third variation uses the prosecution network and the defendant network proposed in this paper, which prunes the jury network.
The comparison results on the MFIRST dataset are shown in Tab. \ref{tab:Ablation-study-on}, where PNet\_v1 represents the dense connection transformer without a fine-grained attention module.
PNet is the prosecution network of CourtNet.
CourtNet\_v1 represents using the prosecution network and the defendant network but without the jury network, the final result is the mean of their outputs. 

The fine-grained attention module can improve the detection performance.
Compared PNet\_v1 with PNet, the detection performance of PNet\_v1 (F1-score: 0.59) is better than PNet (F1-score: 0.55).
The reason is that the fine-grained attention module is performed inside each patch, enabling the network to pay attention to small targets at the pixel level.
Otherwise, the network just pays attention to small targets at the patch level.

The function of the dense connection is to reduce the dimension of the attention map.
According to the principle of tensor addition, the dimension of the input feature must be equal to that of the output feature.
Without the dense connection, the dimension of the attention map of the fine-grained attention module must be $768\times768$, which is far larger than the available GPU memory.
With the dense connection, the dimension of the attention map of the fine-grained attention module is $32\times32$ (the details of the fine-grained attention module can be seen in Sec. \ref{sec:DCT}), so that the implementation of the fine-grained attention module is possible. 

The prosecution-defendant-jury network structure can improve the detection performance.
Compare CourtNet\_v1 with CourtNet, with only a small cost of parameters (0.1M) and flops (0.01GMac), CourtNet can achieve a large improvement (4\% for F1-score), which shows the effectiveness of the prosecution-defendant-jury network structure.

\begin{figure*}[!t]
\begin{centering}
\includegraphics[width=0.99\textwidth]{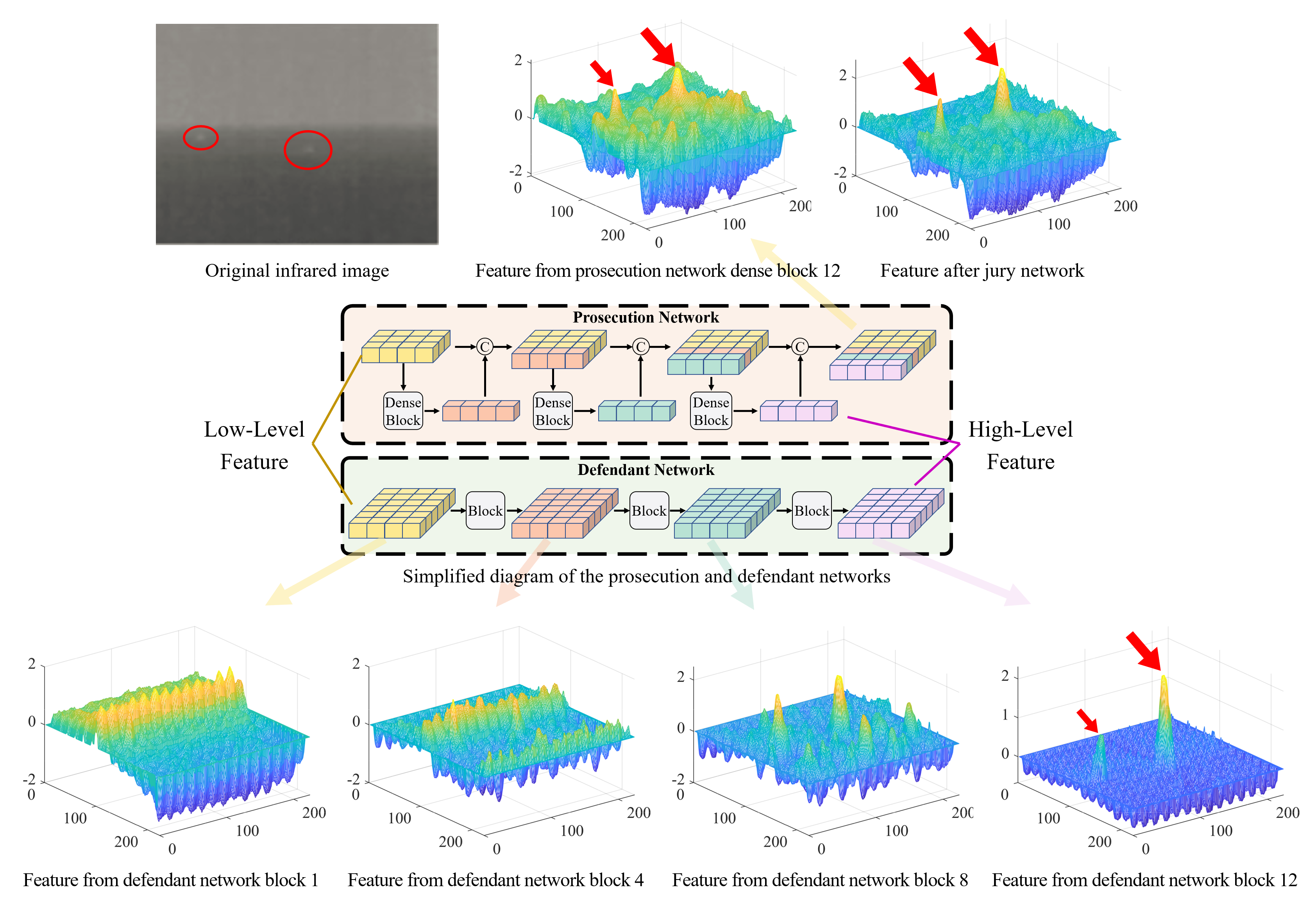}
\end{centering}
\caption{\label{fig:Comparison-of-the-str}Comparison of the structures and features between the prosecution and defendant networks.
The features extracted from the prosecution network contain both low-level and high-level features, which contain original information but much fluctuation.
The low-level features extracted from the defendant network are gradually replaced by high-level features during the forward propagation, in which fluctuation becomes fewer but are prone to lose real targets.
After judging by the jury network, the two targets are both easy to distinguish.}
\end{figure*}

\subsubsection{Adaptive Balance Loss}

The adaptive balance loss can reduce fluctuations and accelerate convergence.
We compared the training process when using the adaptive balance loss and not.
Without the adaptive balance loss, the formulation of loss is
\begin{equation}
\mathcal{L}_{\textrm{w/o abl}}=-\log(Pr)-\log(Re).
\end{equation}
The loss of the training set and the F1-score of the test set are shown in Fig. \ref{fig:Comparison-of-the}.
Fig. \ref{fig:Comparison-of-the} (a) shows that the loss when using the adaptive balance loss has a faster convergence speed and much lower fluctuations than the loss without the adaptive balance loss.
This is because when the precision rate (or the recall rate) is good enough, the adaptive balance loss can make the model adaptively optimize the recall rate (or the precision rate).
But without the adaptive balance loss, even if the precision rate (or the recall rate) is good enough, the model greedily tries to continue to optimize the precision rate (or the recall rate), resulting in the occurrence of fluctuations of the loss value.
As a consequence, the model using the adaptive balance loss can be effectively trained and has a good detection performance, which can be shown in Fig. \ref{fig:Comparison-of-the} (b).

In order to test the impact of different $\gamma $ of the adaptive balance loss on performance, 
we use prosecution network with different $\gamma $ (from 0 to 5) to detect small targets, the results are shown in Tab. \ref{tab:gamma}. 
The results show that the performance is best when $\gamma $ is set to 3.
When $\gamma $ is from 0 to 3, the F1-score increases. 
This is because when $\gamma $ is small, the balancing ability of $(1- ·)\gamma $ is insufficient.
When $\gamma $ is greater than 3, the F1-score decreases, because when $\gamma $ is too large, the value of $(1- ·)\gamma $ fluctuates too much, which is adverse to training.

\subsection{Qualitative Performances}

\subsubsection{Prosecution-Defendant-Jury Structure}

In order to illustrate the detection performance of the prosecution network or the defendant network, as well as whether the jury network can judge their reliability, we select 8 samples from MFIRST for comparison.
Fig. \ref{fig:Qualitative-performances-of} shows the infrared images (the first column), the ground truth (the second column), the detection results of each sub-network (the third column for the prosecution network, the fourth column for the defendant network), and the final results after the judgment by the jury network (the last column).
The numbers under the detection results represent the weights $w_{p}$ and $w_{d}$ in Eq. 1.
In Fig. \ref{fig:Qualitative-performances-of}, the differences in the detection results of the prosecution network, the defendant network, and the jury network are all pointed out, where the red arrow indicates that the targets are detected (Positive), and the red circle means the network does not detect any target in this area (Negative).

The images in Fig. \ref{fig:Qualitative-performances-of} (a) show the prosecution network focuses on the recall rate.
The defendant network fails to recall all of the targets (False Negative), shown in the fourth column of Fig. \ref{fig:Qualitative-performances-of} (a).
However, the prosecution network can detect these targets (the third column of Fig. \ref{fig:Qualitative-performances-of} (a)).
The jury network gives appropriate weights, where the weights of the prosecution network (0.96, 0.93, 0.71, 0.80) are larger than that of the defendant network (0.04, 0.07, 0.29, 0.20).

The images in Fig. \ref{fig:Qualitative-performances-of} (b) show the defendant network focuses on the precision rate.
Parts of the targets detected by the prosecution network are not real (False Positive), resulting in a low precision rate, which can be seen in the third column of Fig. \ref{fig:Qualitative-performances-of} (b).
However, the defendant network can filter these fake targets (the fourth column of Fig. \ref{fig:Qualitative-performances-of} (b)).
The jury network gives appropriate weights, where the weights of the defendant network are all 0.99.
Fig. \ref{fig:Qualitative-performances-of} demonstrates the effectiveness of the prosecution-defendant-jury network structure.

The reason why the prosecution network and defendant networks perform differently is the difference in their network structure.
To this end, we compare the structure of the prosecution and defendant networks and the features extracted from them, as shown in Fig. \ref{fig:Comparison-of-the-str}.
As for the defendant network, low-level features that include the original information of infrared images were replaced by high-level features that include semantic information during the forward propagation. 
While for the prosecution network, low-level and high-level features are all retained. 
Compared with the low-level features, high-level features are with fewer fluctuations. Noise or fake targets are filtered.
However, parts of the real dim and small targets are also filtered.
The features extracted from the prosecution network contain both low-level and high-level features, so the prosecution network can identify all suspected targets.
The features extracted from the defendant network only contain high-level features, so the defendant network avoids considering noise as a target.

\begin{figure}[!h]
\begin{centering}
\includegraphics[width=0.30\textwidth]{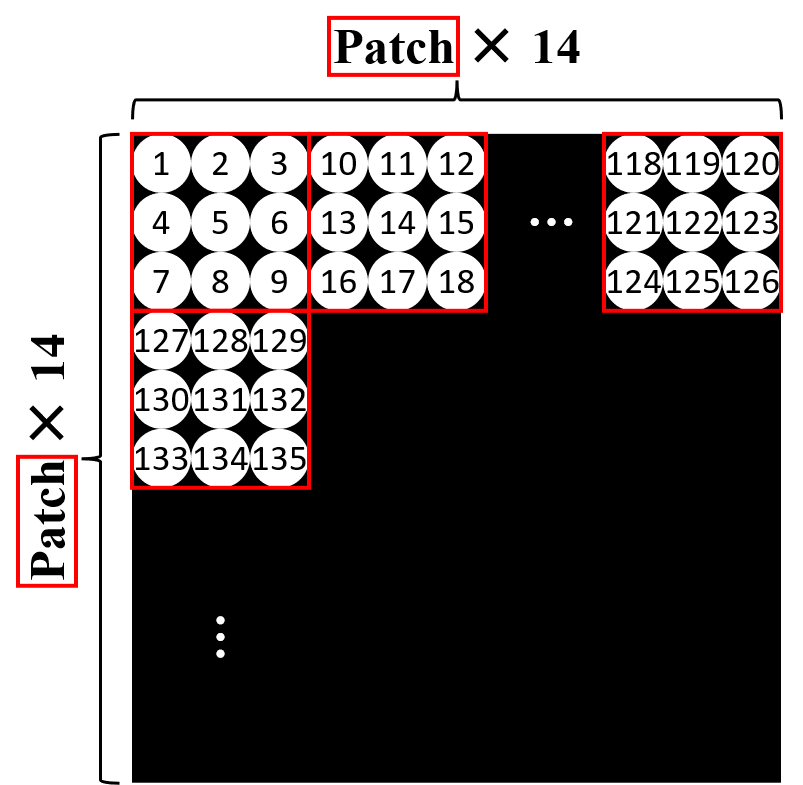}
\par\end{centering}
\caption{\label{fig:order}The order of the target appearance of the constructed dataset.}
\end{figure}

\subsubsection{Coarse-grained Attention vs. Fine-grained Attention}

\begin{figure*}[!h]
\begin{centering}
\subfloat[Targets in the same patch.]{\includegraphics[width=1\textwidth]{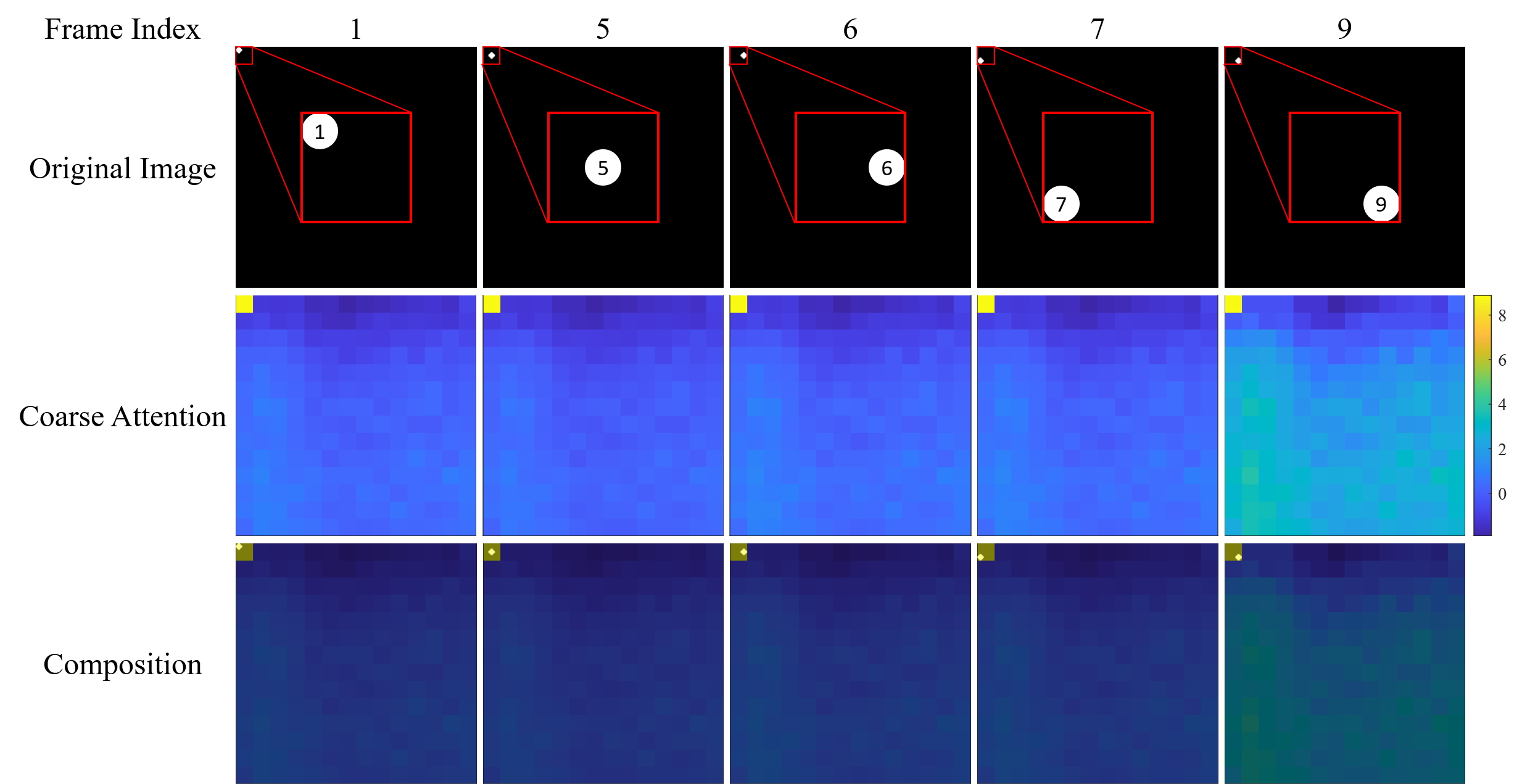}}
\par\end{centering}
\begin{centering}
\subfloat[Targets in different patches.]{\includegraphics[width=1\textwidth]{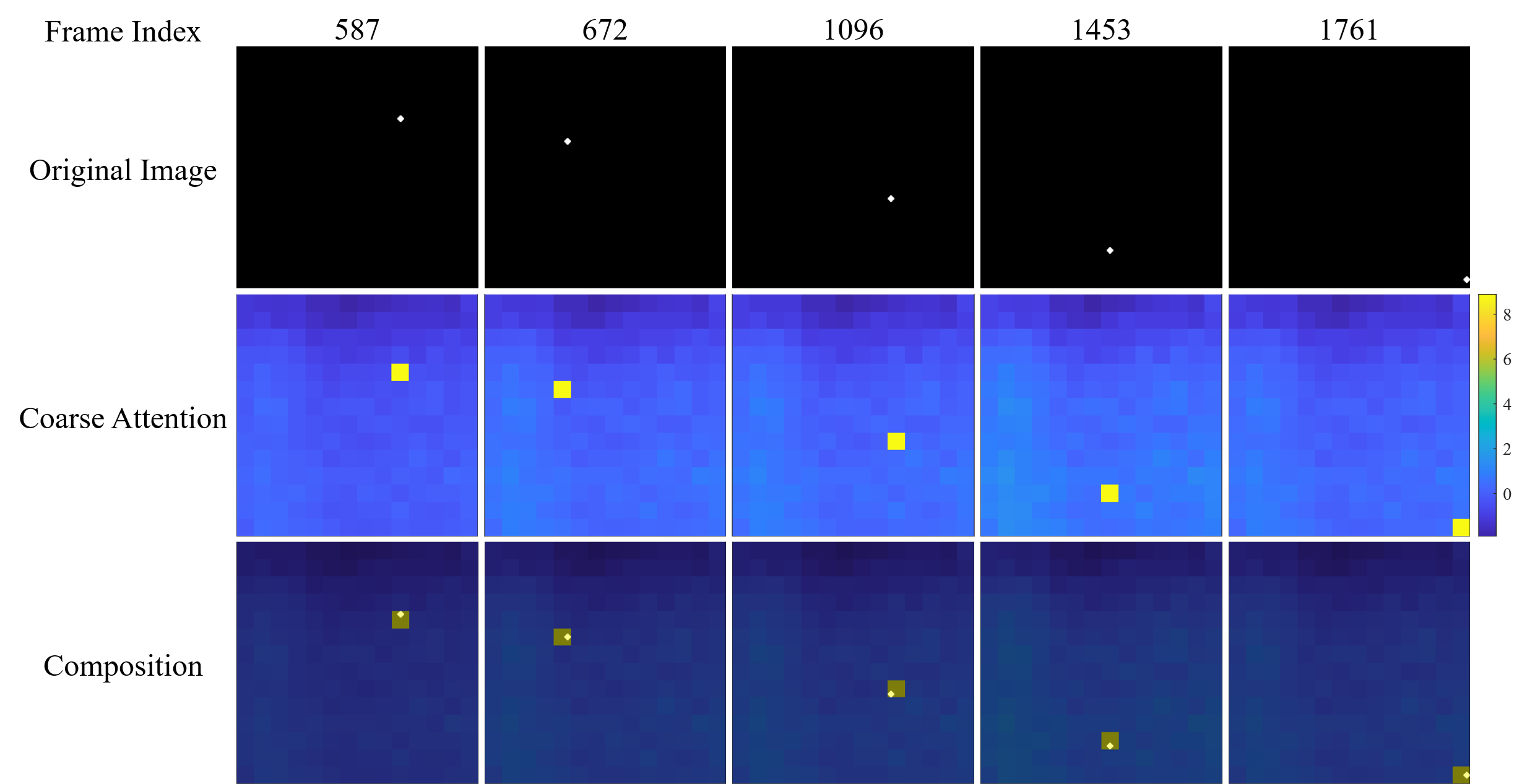}}
\par\end{centering}
\caption{\label{fig:coarse}Qualitative performances of the coarse-grained attention.}
\end{figure*}

In order to verify the effectiveness of the coarse-grained and fine-grained attention modules,
especially whether the coarse-grained attention can locate the small target at the patch level,
and the fine-grained attention can locate the small target inside the patch,
we construct a series of images with small targets.
In our the constructed dataset, there is only one target appears in each image. 
Among them, one image is divided into $14\times 14$ patches (the same as the embedding of Transformers), and patches are arranged row by row. 
Within one patch, there are 9 locations of targets which are also arranged row by row. 
The order of the target appearance can be shown in Fig. \ref{fig:order}.
The constructed dataset can be downloaded at https://github.com/PengJingchao/CourtNet.

We use the prosecution network alone to detect targets. 
All the targets can be detected successfully (detection rate is 100\%, and F1-score is 80\%).
As for the coarse-grained attention, for its features have 196 dimensions, we reshape the 196-dimension features to $14\times 14$ matrixs, 
and displays the data in the matrix as images that use the full range of colors in the colormap, as shown in Fig. \ref{fig:coarse}.
From the figure, the coarse-grained attention can always focus on the patch where targets are located in.
In Fig. \ref{fig:coarse} (a), targets are in the same patch (in the up-left patch of the image) but with different locations.
The coarse-grained attention focuses on the same patch.
In Fig. \ref{fig:coarse} (b), targets are in different patches, and the coarse-grained attention focuses on their corresponding patches.

\begin{figure}[!h]
\begin{centering}
\subfloat[Visualization of part of the features of the fine-grained attention.]{\includegraphics[width=0.45\textwidth]{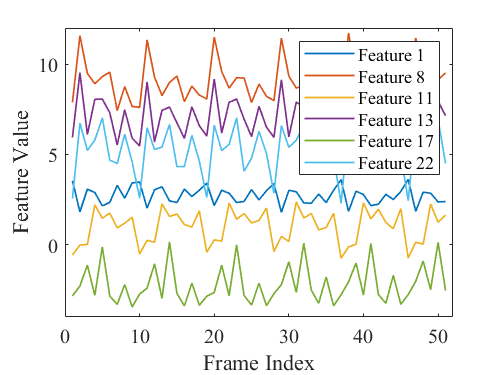}}
\par\end{centering}
\begin{centering}
\subfloat[Power spectrum charts of the features of the fine-grained attention.]{\includegraphics[width=0.45\textwidth]{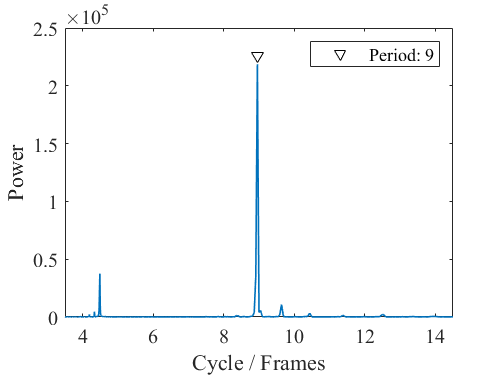}}
\par\end{centering}
\caption{\label{fig:fine}Qualitative performances of the fine-grained attention.}
\end{figure}

As for the fine-grained attention, due to its features having 32 dimensions, we cannot visualize its features like visualization in the coarse-grained attention.
Instead, we arrange its features as a feature series in the the target appearance order.
Since the fine-grained attention is designed to locate the targets inside patches, the same location inside different patches should have similarities for the fine-grained attention.
In other words, the feature series of the fine-grained attention should have cyclicity, because the order of the target location in different patches is the same.
Fig. \ref{fig:fine} (a) visualizes part of the 32-dimension features of the fine-grained attention.
We can see that the feature series indeed have cyclicity.
To figure out the exact period, we analyze 32-dimension features with fast Fourier transform (FFT).
Fig. \ref{fig:fine} (b) plots power as a function of period, measured in frames per cycle.
The plot reveals that the features of the fine-grained attention show similarity once every 9 frames, which is the same as 9 locations inside each patch.
The cyclicity of the feature series of the fine-grained attention demonstrates its effective locating capability inside the patch.

\section{Conclusion}

In this paper, we propose a novel ISTD method, called CourtNet. CourtNet
borrows ideas from court debate and utilizes a prosecution-defendant-jury
network structure. The prosecution network tries to find all suspected
targets to improve the recall rate; the defendant network tries to
avoid regarding noises as targets to increase the precision rate.
The jury network takes their outputs as inputs, judges their reliability,
and weights-sum their outputs to get the final result, so that achieve
adaptively balancing the precision and recall rate. To deal with small
targets, the prosecution network adopts the densely connected transformer
strcture and the fine-grained attention module to accurately locate
small targets. Besides, CourtNet adopts an adaptive balance loss,
which can reduce fluctuations and accelerate convergence. Extensive
experiments on MFIRST and SIRST demonstrate the effectiveness of dealing
with the small size and balancing the precision and recall rate. In
the future, we will try to integrate the model and adopt other advanced
architectures.

\bibliographystyle{plain}
\bibliography{ref}

\end{document}